\documentclass[conference]{IEEEtran}
\IEEEoverridecommandlockouts
\usepackage{cite}
\usepackage{amsmath,amssymb,amsfonts}
\usepackage{algorithmic}
\usepackage{graphicx}
\usepackage{textcomp}
\usepackage{xcolor}
\usepackage{siunitx}
\usepackage{wrapfig}
\usepackage{hyperref}
\usepackage{cleveref}
\usepackage{placeins}
\usepackage{stfloats}

\usepackage[all]{nowidow}

\Crefname{equation}{}{}
\Crefname{figure}{Fig.}{Figs.}
\Crefname{tabular}{Tab.}{Tabs.}

\def\BibTeX{{\rm B\kern-.05em{\sc i\kern-.025em b}\kern-.08em
    T\kern-.1667em\lower.7ex\hbox{E}\kern-.125emX}}

\begin{document}

\title{High-Voltage Optocoupler Amplifier \linebreak for Electrostatic Actuators}

\author{\IEEEauthorblockN{George C. Jurgiel}
\IEEEauthorblockA{\textit{Department of EECS} \\
\textit{Massachusetts Institute of Technology}\\
Cambridge MA, USA}
\and
\IEEEauthorblockN{Alex S. Miller}
\IEEEauthorblockA{\textit{Department of Aeronautics and Astronautics} \\
\textit{Massachusetts Institute of Technology}\\
Cambridge MA, USA}
\and
\IEEEauthorblockN{Jeffrey H. Lang}
\IEEEauthorblockA{\textit{Department of EECS} \\
\textit{Massachusetts Institute of Technology}\\
Cambridge MA, USA}
\thanks{G.J. was funded by the MIT Undergraduate Research Opportunities Program and a Massachusetts Space Grant Fellowship. A.M. was supported by the NSF Graduate Research Fellowship under Grant No. 2141064 and by the Fannie and John Hertz Foundation.}
}

\maketitle

\begin{abstract}
Many electrostatic actuators require multi-kilovolt drive voltages at sub-milliamp currents, a task poorly suited for conventional switching devices. As an alternative, we demonstrate a high-voltage amplifier using optocouplers as active elements. The amplifier produces a 20-kV peak-to-peak output with up to 500\,Hz bandwidth while maintaining a minimal component count. By using optocouplers as linear devices in feedback, lower harmonic distortion and higher bandwidth are achieved than offered by equivalent PWM amplifiers. This design improves the viability of electrostatic actuators by providing a simpler method to achieve useful drive waveforms.
\end{abstract}


\section{Introduction}
Electrostatic actuators continue to be an active area of development as an alternative to conventional magnetic actuators. The thin and light electrodes of electrostatic actuators are a compelling alternative to the large and heavy coils and cores of magnetic devices. Electrostatics are dominant in MEMS devices, but there is also active development of electrostatic actuators at the macro scale. The field of soft robotics uses dielectric elastomer actuators (DEA) \cite{Pelrine} and hydraulically-amplified self-healing electrostatic (HASEL) actuators \cite{Acome}\cite{Mitchell}. There are also emerging electrostatic industrial machines competing more directly with common magnetic machines \cite{Ludois}.

At the MEMS scale, the gaps between electrodes are small, producing sufficient electrostatic forces at low voltage \cite{Neugebauer}. For larger actuators, the gaps grow, requiring higher drive voltages to function well. Their multi-kilovolt amplitudes present challenges not faced when driving more conventional magnetic machines. Electrical devices rated for the kilovolt range are expensive and uncommon. Drive designs often stack devices in series requiring complex gate drivers and isolators\cite{Batra}. This issue extends to circuit construction in which special care is required to prevent breakdown between components. Additionally, electrostatic loads are capacitive, so their energy storage scales with voltage squared. This means the devices must handle significant reactive power \cite{Campolo}.

\begin{figure}[ht!]
    \centering
    \includegraphics[]{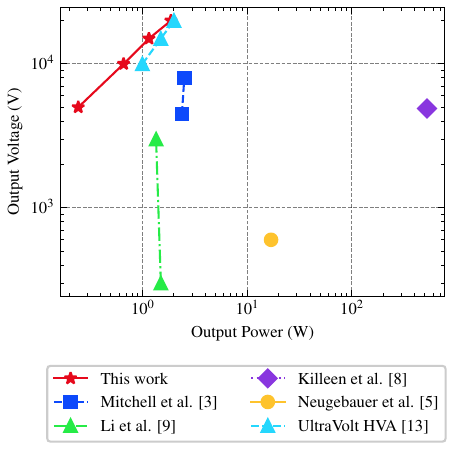}
    \caption{In comparison to other high-voltage amplifiers for electrostatic actuators, the amplifier in this work achieves state-of-the-art voltage at a low component count, while offering true analog waveform control.}
    \label{fig:comparison}
\end{figure}

A general purpose amplifier for driving a variety of electrostatic machines should produce waveforms with amplitudes on the order of \SI{10}{\kilo\volt} to drive machines with large gaps. The amplifier should also operate at frequencies down to DC for capacitive actuators and up into the kilohertz range for use with electrostatic induction machines. 

Previous high-voltage amplifier designs have used high-voltage FETs in an H-bridge \cite{Killeen} and switched-capacitor topologies \cite{Li} to achieve high-voltage waveforms. However, these approaches are limited by component ratings and introduce large component counts with significant design complexity. The unconventional approach of using optocouplers as active elements reduces component count by using devices with higher individual voltage ratings, and simplifies circuit control through optical isolation. Optocouplers have been used before for amplification \cite{Mitchell}\cite{Wiranata}. But, this prior work used optocouplers driven in a PWM mode, requiring a microcontroller for feedback which adds additional complexity and reduces potential bandwidth \cite{Mitchell}\cite{Wiranata}. 

\begin{figure*}[bp]
    \centering
    \includegraphics[width=0.9\textwidth]{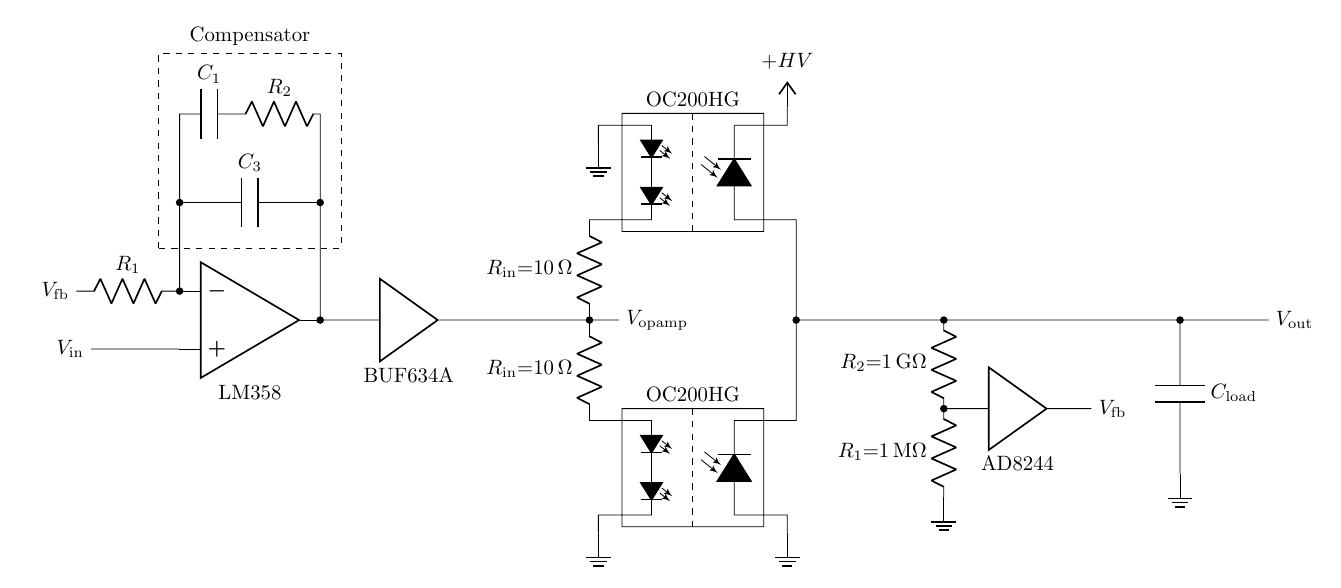}
    \caption{Amplifier uses optocouplers in feedback to achieve low component count for driving capacitive loads at high voltage.}
    \label{fig:amplifier_schematic}
\end{figure*}

Here, we demonstrate a linear amplifier using optocouplers as the main active components, with feedback. While analog systems similar to this work have been proposed\cite{HVOpAmp} no design information, component selection, or experimental data for such designs are publicly available. 

This work is compared to previous designs on the metrics of output voltage and output power in \Cref{fig:comparison}. In addition to high performance by these metrics, we use high-voltage optocouplers, achieving higher voltages than possible with single FETs while maintaining a low component count. Using optocouplers as linear devices allows the converter to have a lower total harmonic distortion and a higher bandwidth than previous similar work using a PWM approach \cite{Mitchell}.

\section{Amplifier Design}
\label{sec:circuit_design}

The amplifier design, shown in \Cref{fig:amplifier_schematic}, uses a pair of  optocouplers (OC200HG\cite{OC200HG}) in a push-pull configuration that operates as a class-B amplifier. The amplifier is designed with analog feedback to remove distortion and correct phase shifts caused by capacitive loading.

\nocite{UltraVolt}

The amplifier operates by modulating the reverse current of each photodiode in a push-pull configuration. The built-in optical isolation of the optocouplers makes them simple to drive whereas a FET would require additional isolators on the gate side capable of floating to the 20-kV rail. Our implementation uses an additional amplifier (BUF634A) after the opamp (LM358) output to drive the optocoupler, but this could be avoided by using an opamp with higher output-current.

Since the optocouplers operate as current sources, the amplifier acts as an integrator when driving capacitive loads, causing a phase shift and attenuation with increasing frequency. Additionally, the LED inputs of the optocouplers are inherently nonlinear, resulting in crossover distortion when driven in a class-B configuration. Contributions from both of these effects render the amplifier unable to operate in an open-loop configuration. By operating the pair of optocouplers in feedback, the controller can correct for the phase shift and high-frequency attenuation, ensuring the desired magnitude and phase. The controller also slews through the crossover distortion region, ensuring an output with low harmonic distortion.

The feedback path used in our implementation is a resistor divider, yielding a reliable design with a low component count. The divider was designed with very high resistances to ensure that it does not degrade output performance. A high-impedance buffer (AD8244) was included so the feedback network and output measurement do not load the divider. This can be avoided by accounting for loading effects in the divider and feedback network design. The resistor divider eliminates optical isolation between the low-voltage input and the high-voltage output. If the low-side resistor fails open, damage to the low-voltage circuitry could result. Nonetheless, the resistor divider was preferred here for voltage sensing because transmitting a high-voltage analog signal across the isolation barrier requires a significant increase in complexity and component count.

\section{Optocoupler Modeling}
\label{sec:opto_modeling}

\begin{figure}[t!]
\includegraphics[width=0.5\textwidth]{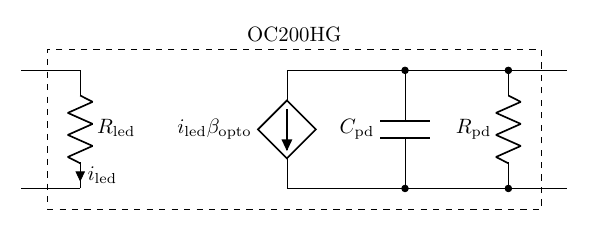}
    \caption{Optocoupler model used to develop feedback circuit.}
    \label{fig:linear_opto_model}
\end{figure}

This amplifier uses commercially-available OC200HG optocouplers which are rated for \SI{20}{\kilo\volt} and \SI{500}{\micro\ampere} output. With the photodiode reverse biased and the LED forward biased, the optocoupler can be modeled as a current-dependent current source with output parasitics as shown in \Cref{fig:linear_opto_model}.

\begin{figure}[t!]
    \centering
    \includegraphics[]{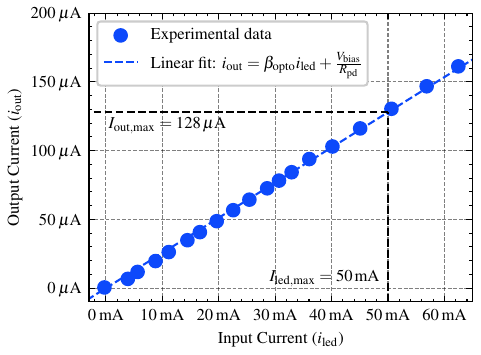}
    \caption{Current input to output relationship for the OC200HG optocouplers showing a current gain of $\beta = 0.00256$ which is independent of output bias.}
    \label{fig:optocoupler_current}
\end{figure}

The DC current gain and resistance of the optocoupler were determined by measuring the output current over a range of input currents as shown in \Cref{fig:optocoupler_current}. The slope of the relation was used to determine the optocoupler gain $\beta_{\mathrm{opto}}=0.00256$. The y-intercept was used to find the equivalent output resistance of $R_{\mathrm{pd}}=\SI{5}{\mega\ohm}$. During this process, the equivalent small-signal LED resistance was found to be $R_{\mathrm{led}}=\SI{3}{\ohm}$ by measuring the I-V characteristics of the input. Being a diode, the input LED has a highly nonlinear I-V relationship. Linearizing this relationship to be the constant $R_\mathrm{led}$ shown in \Cref{fig:linear_opto_model} enables an accurate analysis for feedback design and small-signal behavior. This testing was performed with a DC bias of \SI{10}{\kilo\volt} on the photodiode, although the current gain is independent of bias.

\begin{figure}[t!]
    \centering
    \includegraphics[width=0.5\textwidth]{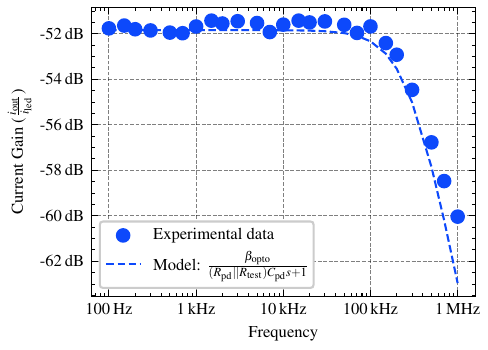}
    \caption{Frequency response of optocoupler gain using a \SI{100}{\kilo\ohm} test load. The cutoff frequency was used to determine $C_{\mathrm{pd}}$.}
    \label{fig:optocoupler_frequency_response}
\end{figure}

The photodiode has a junction capacitance that limits its bandwidth. This was found to be $C_{\mathrm{pd}}=\SI{5.5}{\pico\farad}$ by measuring the frequency response of optocoupler current gain shown in \Cref{fig:optocoupler_frequency_response}. From the \SI{-3}{\decibel} cutoff frequency, the RC time constant for the circuit was determined and used to find $C_{\mathrm{pd}}$. For this test a load of $R_{\mathrm{test}}=\SI{100}{\kilo\ohm}$ was used which was incorporated into the capacitance calculation.

The OC200HG optocoupler can source current up to \SI{500}{\micro\ampere}. We limited our design to a lower maximum current as set by the \SI{10}{\ohm} input resistor. The opamp output is limited to $\pm$\SI{3.5}{\volt}, and the two series-connected optocoupler LEDs have a combined forward voltage of \SI{3}{\volt}. This leads to a maximum LED current of $I_{\mathrm{led,max}}=\SI{50}{\milli\ampere}$ and a corresponding output current limit of $I_{\mathrm{out,max}}=\SI{128}{\micro\ampere}$ as shown in \Cref{fig:optocoupler_current}. For many low-power machines this is sufficient, but it becomes a limitation for higher-power and/or high-frequency applications where reactive currents can exceed this limit. This results in an output slew rate limit of 
\begin{equation}\label{eq:slew-rate-limit}
    \frac{dV_{\mathrm{out}}}{dt}=\frac{I_{\mathrm{out,max}}}{2C_{\mathrm{pd}}+C_{\mathrm{load}}},
\end{equation}
where $I_{\mathrm{out,max}} =$ \SI{128}{\micro\ampere} is the maximum output current set by $R_{\mathrm{in}}$, and $C_{\mathrm{load}}$ is the load capacitance.  For calculations of slew rate limits and bandwidth, the impedance of $C_{\mathrm{pd}}$ is much smaller than $R_{\mathrm{pd}}$ for the frequency range of interest and therefore $R_{\mathrm{pd}}$ may be neglected. For sinusoidal excitation, the maximum achievable frequency $f_{\mathrm{max}}$ is
\begin{equation}
    \label{eq:sine_bandwidth_limit}
    f_{\mathrm{max}} = \frac{I_{\mathrm{out,max}}}{\pi V_{\mathrm{out,pp}} (2C_{\mathrm{pd}} + C_{\mathrm{load}})},
\end{equation}
where $V_{\mathrm{out,pp}}$ is the peak-to-peak output voltage. If the frequency is pushed above this, the slew rate limit will distort the waveform into a triangle wave and cause attenuation. Under these conditions, the \SI{-3}{\decibel} cutoff frequency is

\begin{equation}
    \label{eq:3db_bandwidth_limit}
    f_{\mathrm{\SI{-3}{\decibel}}} = \frac{I_{\mathrm{out,max}}}{\sqrt2V_{\mathrm{out,pp}} (2C_{\mathrm{pd}}+C_{\mathrm{load}})}.
\end{equation}

\section{Feedback Design}
\label{sec:feedback}

Electrostatic actuators require control over their drive voltages to manage force, but optocouplers provide control over current. Since the amplifier must drive varying loads, feedback is required to maintain the desired output voltage. Feedback also removes crossover distortion caused by the push-pull configuration of the optocouplers. The system is simple enough that analog feedback is appropriate, avoiding the need for microprocessors or digital signal processing.

Eliminating instability caused by driving the capacitive loads presented by electrostatic actuators requires feedback compensation. Without compensation, configuring the amplifier in direct feedback with an opamp creates an amplifier with low stability. When driving a capacitive load, the push-pull amplifier transfer function derived from the model in \Cref{fig:linear_opto_model} is
\begin{equation}
    \label{eq:opto_gain}
    G_{\mathrm{amp}}(s) = \left( \frac{\beta_{\mathrm{opto}} }{R_{\mathrm{in}}+R_{\mathrm{led}}} \right) 
    \left( \frac{R_{\mathrm{pd}}}{1+sR_{\mathrm{pd}}(2C_{\mathrm{pd}}+C_{\mathrm{load}})}\right).
\end{equation}
At the crossover frequency of the amplifier, this transfer function has a phase shift of \SI{90}{\degree} in the closed-loop return ratio. Opamps have a near-DC pole, adding an additional \SI{90}{\degree} phase shift in the return ratio, which would result in a loop having a low phase margin and low stability. Additionally, without compensation, low stability would result in distortion for all excitations due to crossover distortion. Crossover distortion leads to a quasi-square-wave intermediate voltage in the feedback loop. Without compensation, each step of this intermediate voltage would lead to poorly damped oscillations near the return ratio crossover frequency when there is low stability, which propagate to the output distorting the waveform.

To increase stability and reduce distortion, lead compensation is required to boost the return-ratio phase margin at the crossover frequency. This can be implemented using a Type-II compensator\cite{Lee} of the form
\begin{equation}
    G_{\mathrm{comp}}(s)=\frac{1+C_1R_2s}{(C_1+C_3)R_1s+C_1C_3R_1R_2s^2}\ .
\end{equation}
This compensator adds a phase boost of $\phi_{\mathrm{peak}} = \sin^{-1}\left(\frac{\alpha-1}{\alpha+1}\right)$ at the frequency $f_{\mathrm{\phi peak}} = \frac{1}{\tau\sqrt{\alpha}}$, where $\tau = \frac{C_1 C_3 R_2}{C_1+C_3}$ and $\alpha = \frac{C_1+C_2}{C_3}$.

Since the amplifier frequency is slew-rate limited well below the crossover frequency, a large phase margin does not degrade the performance of the amplifier. For this design, a crossover frequency of \SI{50}{\kilo\hertz} and phase margin of \SI{85}{\degree} were targeted using component values $R_1=\SI{1}{\kilo\ohm}$, $R_2=\SI{200}{\ohm}$, $C_1=\SI{1}{\nano\farad}$, and $C_3=\SI{470}{\nano\farad}$.

\section{Implementation}

\begin{figure}[t!]
    \centering
    \includegraphics[width=0.5\textwidth]{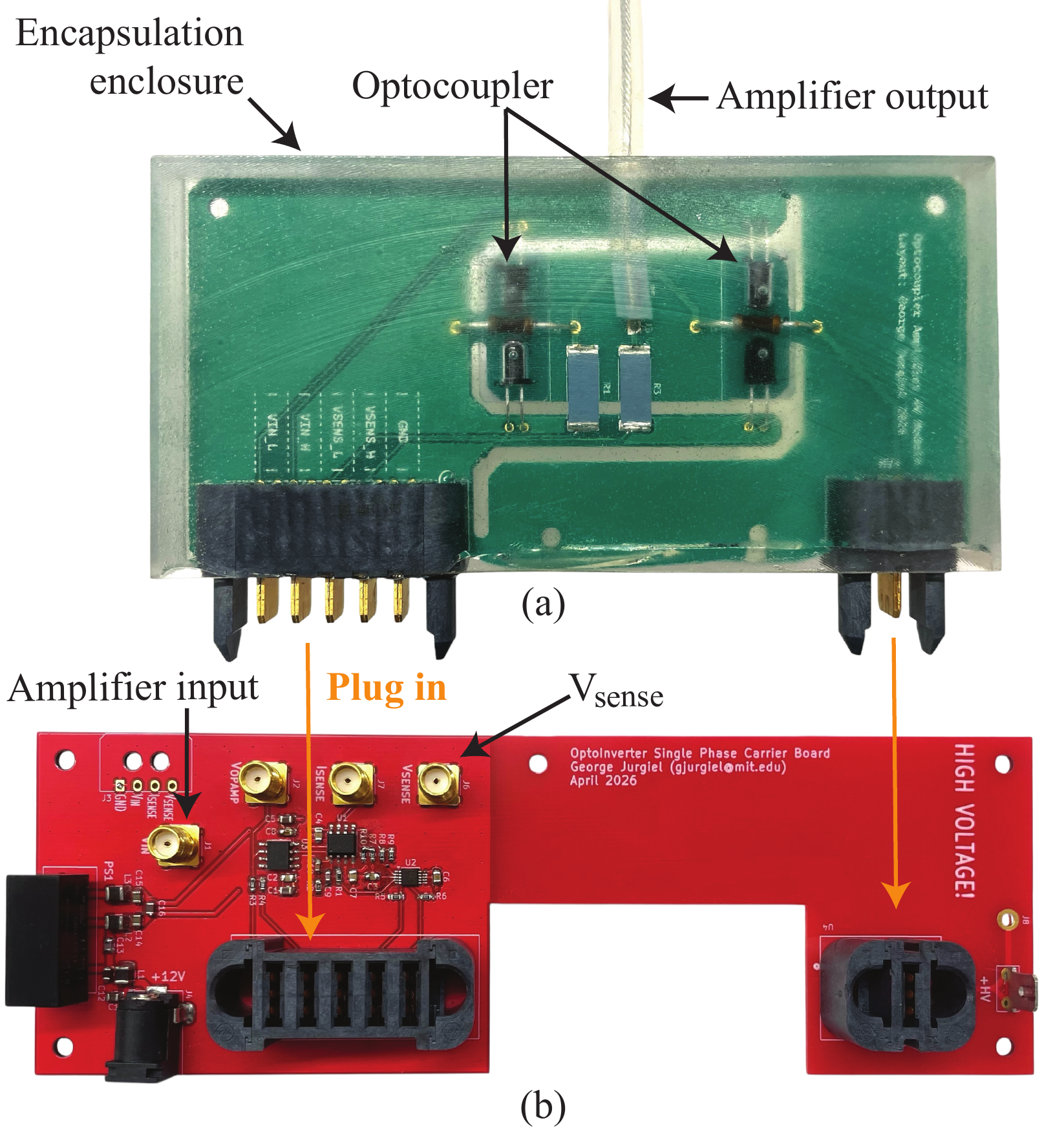}
    \caption{High-voltage amplifier implementation as tested in this paper with (a) encapsulated high-voltage ``module'' and (b) low-voltage carrier board.}
    \label{fig:pcb}
\end{figure}

The amplifier was implemented using two PCBs to separate high-voltage and low-voltage components. The prototype amplifier tested in this paper is shown in \Cref{fig:pcb}. The high-voltage components were encapsulated using clear epoxy resin (MG Chemicals 832WC) in a 3D-printed enclosure to prevent breakdown. This leaves the low-voltage components free for repair and modification, and allows the same high-voltage ``module'' to be reused with different low-voltage control boards, or for a damaged high-voltage ``module'' to be replaced without assembling and encapsulating a new amplifier. This modular replaceability is particularly useful for multichannel and poly-phase amplifiers. The design was built intentionally over-sized to minimize high-voltage creep and interference for improved measurements. The high-voltage output exits the encapsulated enclosure via high-voltage insulating hookup wire (Cicoil Hi-Flex 20AWG). Similar previous work shows a higher density of components \cite{Mitchell} than the implementation shown here. Such high density could also be achieved here with layout modification. All test data presented in this paper were taken using the encapsulated module shown in \Cref{fig:pcb}.

\begin{figure}[h!]
    \centering
    \includegraphics[width=0.5\textwidth]{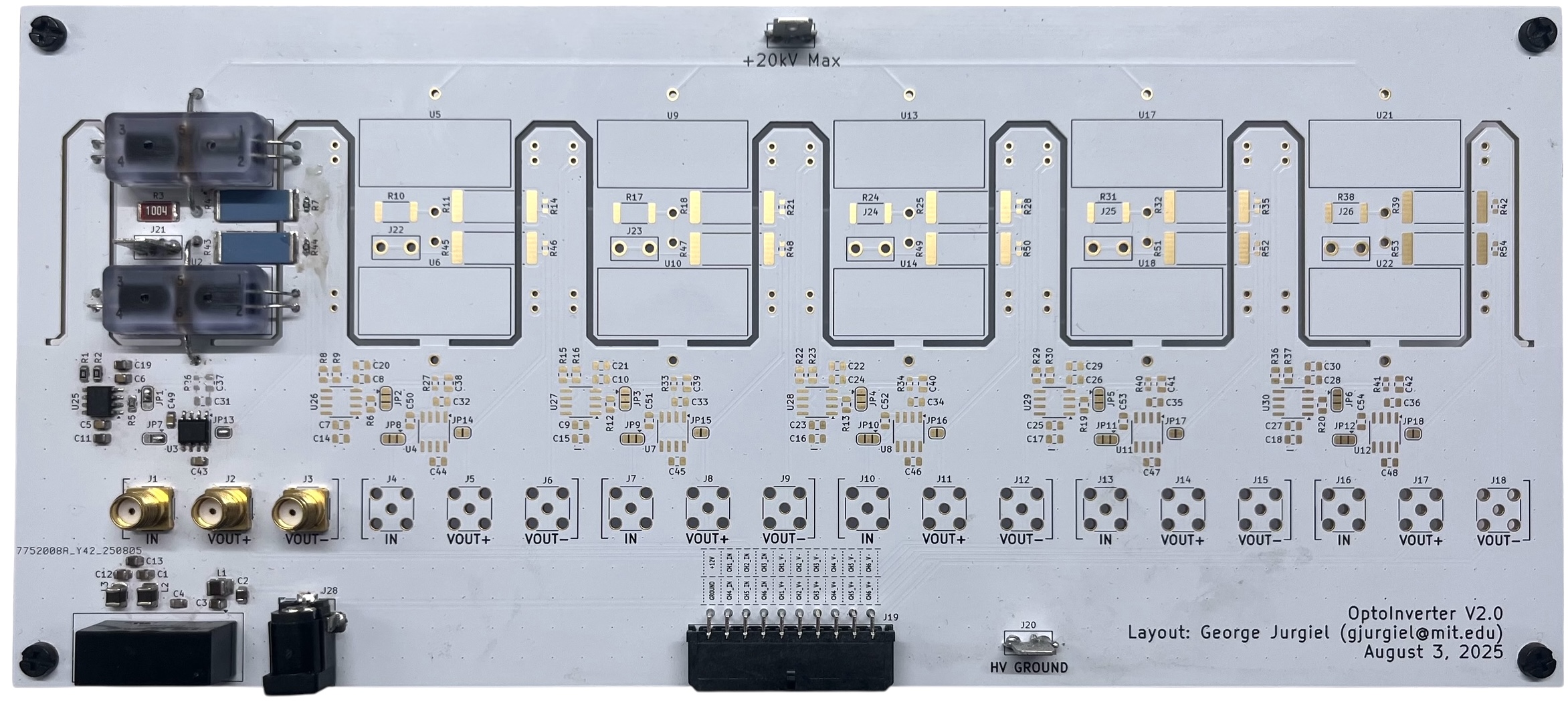}
    \caption{Alternative implementation with populated amplifier on six-channel PCB without buffer components or encapsulated ``module''.}
    \label{fig:pcb_alternative}
\end{figure}

An alternative implementation having six phases on a single PCB without encapsulation was also explored as pictured in \Cref{fig:pcb_alternative}. This design omitted buffer stages in the low-voltage feedback and sensing lines to achieve a smaller component count. This alternative implementation achieved voltages up to \SI{10}{\kilo\volt}, with performance similar to the design tested in this paper. Due to lack of encapsulation in this implementation, this design experienced dielectric breakdown at voltages above \SI{10}{\kilo\volt}.

\section{Results}
\label{sec:results}

\begin{figure*}[t!]
    \includegraphics[width=\linewidth]{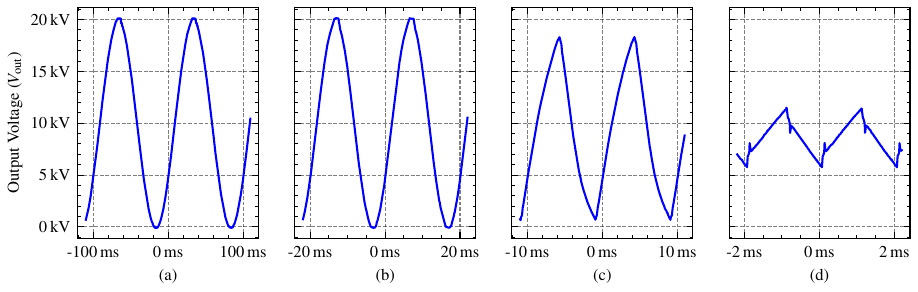}
    \caption{Experimental data of a \SI{20}{\kilo\volt} target output sine wave driving a \SI{33}{\pico\farad} load with frequency of (a) \SI{10}{\hertz}, (b) \SI{50}{\hertz}, (c) \SI{100}{\hertz} and (d) \SI{500}{\hertz}. The amplifier produces low-distortion waveforms at low frequencies that become slew-rate limited leading to distortion and amplitude reduction at high frequencies.}
    \label{fig:time_domain_progression}
\end{figure*}

We collected experimental data for the amplifier shown in \Cref{fig:pcb} at frequencies from DC to \SI{100}{\kilo\hertz}, drive amplitudes from \SI{5}{\kilo\volt} to \SI{20}{\kilo\volt}, and capacitive loads from \SI{10}{\pico\farad} to \SI{100}{\pico\farad}. These conditions are characteristic of those needed to drive electrostatic machines. An example of time-domain waveforms generated by the amplifier at \SI{20}{\kilo\volt}, driving a \SI{33}{\pico\farad} load  at \SI{10}{\hertz}, \SI{50}{\hertz}, \SI{100}{\hertz} and \SI{500}{\hertz} is shown in \Cref{fig:time_domain_progression}. The slew rate limit described in \Cref{eq:slew-rate-limit} results in sinusoidal output waveforms that distort into a triangle wave at higher frequencies. At \SI{10}{\hertz} and \SI{50}{\hertz} the waveform are sinusoidal. At \SI{100}{\hertz} the peaks and troughs begin to distort, and at \SI{500}{\hertz} the waveform becomes fully triangular and significantly attenuated.

\begin{figure}[ht!]
    \centering
    \includegraphics[width=\linewidth]{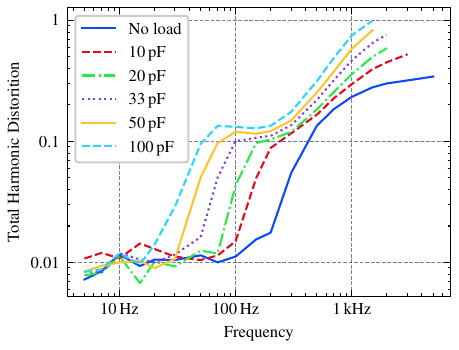}
    \caption{Total harmonic distortion of output waveform for various frequencies and capacitive loads at targeted output amplitude of \SI{20}{\kilo\volt} peak-to-peak.}
    \label{fig:total_harmonic_distortion}
\end{figure}

The relationship between distortion and frequency is shown quantitatively in \Cref{fig:total_harmonic_distortion} for a targeted excitation of \SI{20}{\kilo\volt} and various capacitive loads. A higher capacitive load yields a higher total harmonic distortion (THD) because it reduces the slew rate limit. High drive amplitudes similarly cause the output to distort at lower frequencies due to the slew rate limit, although this is not shown in the figure. At frequencies below \SI{100}{\hertz} the amplifier has a total harmonic distortion of 1\% which arises primarily from measurement noise. As frequency increases and the slew rate limit distortion dominates, the THD rises to 12\% characteristic of a triangle wave. Upon increasing frequency further, other distortion artifacts begin to dominate as the signal becomes attenuated increasing THD further.

\begin{figure}[ht!]
    \includegraphics[width=\linewidth]{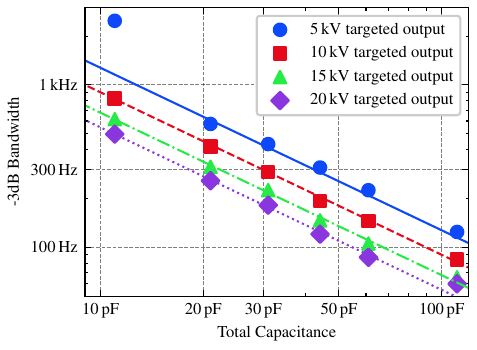}
    \caption{\SI{-3}{\decibel} output bandwidth for various total capacitance ($C_{\mathrm{pd}} + C_{\mathrm{load}}$) and targeted drive voltages.}
    \label{fig:load_bandwidth_plot}
\end{figure}

\begin{figure}[ht!]
    \includegraphics[width=\linewidth]{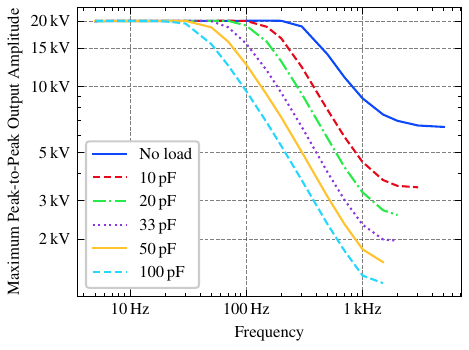}
    \caption{Amplifier frequency response with various capacitive loads for a targeted peak-to-peak amplitude of \SI{20}{\kilo\volt}.}
  \label{fig:20kV_frequency_response}
\end{figure}

\begin{figure}[ht!]
    \includegraphics[width=\linewidth]{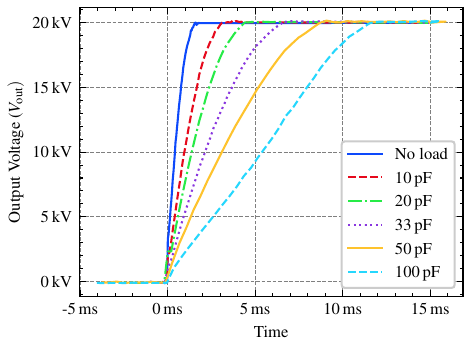}
    \caption{Amplifier step response with various capacitive loads, showing slew rate limit behavior.}
    \label{fig:20kV_step_response}
\end{figure}

We experimentally validated the frequency limit of \Cref{eq:3db_bandwidth_limit} in \Cref{fig:load_bandwidth_plot} by plotting the frequency at which the signal is attenuated by \SI{-3}{\decibel} for various targeted drive amplitudes and loads. The plotted capacitances include parasitics. Attempting to drive the amplifier at a frequency beyond the limit of \Cref{eq:3db_bandwidth_limit} produces an attenuated waveform. This attenuation can be seen in the large-signal frequency response of \Cref{fig:20kV_frequency_response}. The attenuation follows the expected \SI{-20}{\decibel / decade} fall-off with greater attenuation for higher capacitance. The factor that limits frequency and creates distortion is the slew rate limit described by \Cref{eq:slew-rate-limit}. This is shown experimentally in \Cref{fig:20kV_step_response}, the large-signal step response of the amplifier. The figure shows the response for a full 20-kV step but the slew rate is independent of step size.

\begin{figure}[ht!]
    \centering
    \includegraphics[width=\linewidth]{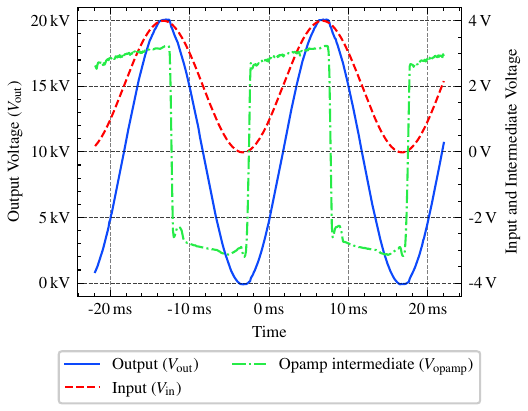}
    \caption{Amplifier output with quasi-square wave intermediate voltage due to crossover distortion.}
    \label{fig:crossover_distortion}
\end{figure}

The necessity of feedback for this amplifier can be observed in the intermediate waveform of \Cref{fig:crossover_distortion}. The opamp output approximates a square wave as it slews through the dead zone of the LED forward voltage. The square wave also leads the output voltage to correct for the phase shift of driving a capacitive load.

\begin{figure}[ht!]
    \centering
    \includegraphics[width=\linewidth]{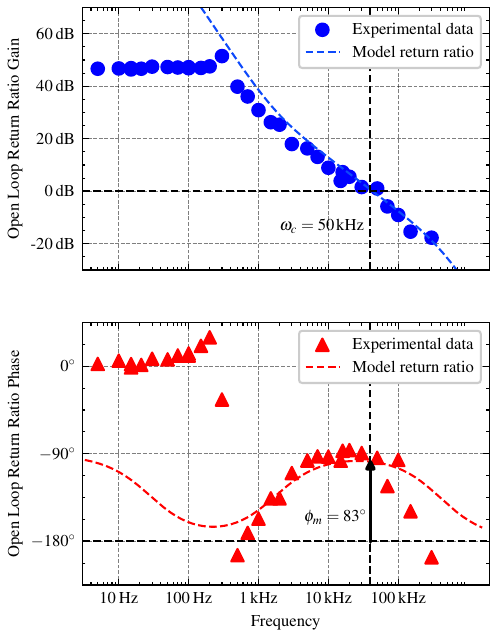}
    \caption{Small signal open loop return ratio plot for the amplifier showing the phase boost provided by the compensator network at the crossover frequency to prevent ringing.}
    \label{fig:small_signal_return_ratio}
\end{figure}

A more detailed understanding of the amplifier feedback can be derived from the small-signal open-loop return ratio of the amplifier shown in \Cref{fig:small_signal_return_ratio}. The amplifier has a small signal crossover frequency of \SI{50}{\kilo\hertz} and a phase margin of \SI{83}{\degree}. The prominent boost in phase near crossover is created by the compensator and ensures that the amplifier is stable. Without compensation, the phase boost would not exist, and the amplifier would have a phase margin near zero. This stability also appears as a lack of ringing in the intermediate voltage steps in \Cref{fig:crossover_distortion}.

The experimental data match the return-ratio model well around the crossover frequency. Deviations at low frequencies exist due to the limited open-loop gain of the opamp not accounted for in the model. However, since the region near the crossover frequency dictates the closed-loop response, this deviation is inconsequential. 

The closed-loop bandwidth is approximately equivalent to the crossover frequency at \SI{50}{\kilo\hertz}. This small-signal bandwidth characterizes how the amplifier responds to small changes in load. The amplifier is not able to drive a full-voltage waveform at this frequency since it is much higher than the large-signal bandwidth of \Cref{fig:20kV_frequency_response}, but this higher small-signal bandwidth is useful for quickly making small adjustments; for instance, understanding this closed-loop bandwidth is useful for the closed-loop operation of actuators.

\section{Conclusion}
\label{sec:conclusion}

We use optocouplers to create an amplifier with a \SI{20}{\kilo\volt} voltage capability, a \SI{128}{\milli\ampere} current capability, and up to a \SI{500}{\hertz} bandwidth, appropriate for driving electrostatic actuators. The amplifier improves upon previous optocoupler-based designs by driving the optocouplers linearly in feedback. This allows for improved bandwidth and a lower component count than designs driving the optocouplers in a PWM mode \cite{Mitchell}. This makes the amplifier compelling for both actuators operating near DC, such as HASEL actuators, and for electrostatic devices that require AC waveforms, such as motors.

Future work may be able to achieve further improvements to amplifier capability. First, for the design implemented in this paper, the optocouplers were conservatively driven up to $i_{\mathrm{led}}=$ \SI{50}{\milli\ampere}, leading to an output capability of \SI{128}{\milli\ampere}. If the optocoupler LEDs were driven near the rated $i_{\mathrm{led}}=$ \SI{100}{\milli\ampere}, then the maximum output current could be increased to \SI{256}{\milli\ampere}. Second, parasitic capacitances in the system also reduce the bandwidth of the existing amplifier, and efforts to minimize parasitics through PCB design, output wire selection, and even photodiode design would improve bandwidth. Third, for multi-channel applications, the form factor of the design could be significantly reduced to that of \cite{Mitchell}, in a fully-encapsulated form. Fourth, for some electrostatic loads, an amplifier with both positive and negative voltage output capability may be useful. The current design can accommodate this, with minor board layout changes to provide sufficient clearance between low voltage control circuitry and the both sides of the high voltage bus. Fifth, additional designs featuring parallel optocouplers for higher current capability could be implemented using high-voltage ``modules'' identical to the current design, and a dual-mounted carrier board. Designs featuring series optocouplers for higher voltage capability could also be implemented, but would require additional circuitry to ensure that none of the devices goes above its voltage rating.

The amplifier presented in this paper is well suited for research on low-power, high-voltage, electrostatic actuators. It is capable of producing a wide variety of excitations useful for testing such actuators. The amplifier also achieves a low component count using only off-the-shelf parts, with an understandable topology allowing for future replication. 

\section{Data and File Availability}

Data and design files are posted at https://github.com/gjurgiel/optocoupler-amplifier

\section{Acknowledgments}
The authors thank the T.J. Rodgers RLE Laboratory for their helpful equipment and advice. 

\bibliographystyle{IEEEtran}
\bibliography{Bibliography}
\end{document}